\documentclass[11pt]{article}
 
\pdfoutput=1

\usepackage[T1]{fontenc}
\usepackage[utf8]{inputenc}
\usepackage{times}
\usepackage{latexsym}
\usepackage{inconsolata}
\usepackage[final]{ACL2023} 

\usepackage{microtype}

\usepackage{amsmath}
\usepackage{amssymb}
\usepackage{amsthm}

\usepackage{algorithm}
\usepackage{algpseudocode}

\usepackage{booktabs}
\usepackage{multirow}
\usepackage{graphicx} 
\usepackage{graphicx}
\usepackage{tikz}
\usepackage{tabularx} 

\usepackage{xcolor}

\usepackage{hyperref}

\usepackage{natbib}

\usepackage{xcolor}
\usepackage{array} 
\usepackage{tabularx} 
\usepackage{multirow}  
\usepackage{float}  
\usepackage{fancyvrb}
\usepackage{tabularx}  
\usepackage{threeparttable}
\usepackage{makecell}
\usepackage{booktabs}
\usepackage{adjustbox}

\usepackage{amsmath}

\usepackage{listings}
\usepackage{xcolor}

\title{FlowPlan-G2P: A Structured Generation Framework\\ for Transforming Scientific Papers into Patent Descriptions}

\author{
  Kris W Pan$^*$ \\
  Amazon \\
  \texttt{kriswpan@amazon.com} \\\And
  Yongmin Yoo$^*$ \\
  Macquarie University \\
  \texttt{yooyongmin91@gmail.com} \\
  \AND
  \normalfont $^*$\textit{Equal contribution.} \\
  \normalfont \textit{Authors are listed in alphabetical order.}
}

\begin{document}
\maketitle
\begin{abstract}
Researchers routinely publish inventions as scientific papers, yet turning them into patent descriptions remains difficult because statutory enablement demands rigid, legally structured disclosure. Prior patent drafting methods either assume patent-native inputs such as claims or generate sections from outline chunks via local retrieval without an explicit document-level representation of technical dependencies. We propose \textsc{FlowPlan-G2P}, a graph-mediated framework that (1) induces a directed concept graph of invention elements and functional relations, (2) partitions it into ordered section subgraphs scored by connectivity, homogeneity, and domain-constraint penalties, and (3) synthesizes patent descriptions using section-specific drafting strategies. On Pap2Pat-EvalGold, Zero-Shot achieves the highest BERTScore at 0.87 yet receives only 1.5 out of 5 on expert legal compliance compared to 4.7 for \textsc{FlowPlan-G2P}, where lexical overlap and expert ratings rank systems in opposite order. Under expert-aligned Pat-DEVAL evaluation, open-weight Llama-4-Scout with \textsc{FlowPlan-G2P} reaches an overall score of 4.3 compared to 2.3 for Few-Shot Sonnet-4.5. We release our code publicly at \url{https://github.com/repoofideas/flowplan_g2p}.\looseness=-1
\end{abstract}
\section{Introduction}

In 2023, 3.55 million patent applications were filed worldwide~\citep{wipo2024report}, yet drafting patent specifications remains slow, costly, and dependent on specialized legal expertise~\citep{jiang2024nlpinsurveys}. Researchers are accustomed to reporting new inventions in scientific papers, but converting a paper into a patent description poses a distinct barrier. Whereas academic papers persuade peers through concise hypotheses and empirical results, the description forms the bulk of a patent specification and must satisfy statutory standards of enablement and sufficiency of disclosure. Meeting these standards requires exhaustive implementation detail, rigid section conventions, and consistent cross-section references across embodiments, creating a sharp structural and linguistic gap from academic writing~\citep{Althammer2021LinguisticallyInformedMasking}.

\begin{figure}[t]
\centering
\begingroup

\definecolor{badred}{HTML}{BF6262}
\definecolor{goodgreen}{HTML}{55866A}

\resizebox{\columnwidth}{!}{%
\begin{tikzpicture}[
    font=\normalsize,
    heading/.style={
        font=\bfseries\normalsize,
        text=black!88,
        align=center,
        inner sep=0pt
    },
    tinycap/.style={
        font=\footnotesize,
        text=black!60,
        align=center,
        inner sep=1pt
    },
    doc/.style={
        rounded corners=3pt,
        draw=black!50,
        fill=white,
        line width=.8pt
    },
    textbar/.style={
        draw=black!32,
        line width=1.3pt,
        line cap=round
    },
    flow/.style={
        ->,
        >=stealth,
        draw=black!48,
        line width=.85pt
    },
    cnode/.style={
        circle,
        draw=black!62,
        fill=white,
        minimum size=.44cm,
        inner sep=0pt,
        outer sep=0pt,
        line width=.85pt,
        font=\scriptsize\bfseries,
        text=black!88
    },
    snode/.style={
        circle,
        draw=black!55,
        fill=white,
        minimum size=.36cm,
        inner sep=0pt,
        outer sep=0pt,
        line width=.8pt,
        font=\scriptsize\bfseries,
        text=black!85
    },
    pill/.style={
        rounded corners=3pt,
        minimum width=1.88cm,
        minimum height=.82cm,
        align=center,
        inner sep=2pt,
        line width=.75pt
    }
]

% Canvas
\path[use as bounding box] (-0.10,-0.05) rectangle (9.50,4.55);

% Panel headings & divider
\node[heading] at (2.20,4.28) {(a) Flat / chunked};
\node[heading] at (7.10,4.28) {(b) \textsc{FlowPlan-G2P}};
\draw[black!15, line width=.65pt] (4.58,0.15) -- (4.58,4.35);

% ============================================================
% Panel (a): Paper -> fragmented draft
% ============================================================
\node[tinycap] at (0.93,3.82) {Paper};
\node[tinycap] at (3.47,3.82) {Draft};

\draw[doc] (0.28,1.98) rectangle (1.58,3.60);
\draw[textbar] (0.48,3.32) -- (1.38,3.32);
\draw[textbar] (0.48,3.06) -- (1.28,3.06);
\draw[textbar] (0.48,2.80) -- (1.40,2.80);
\draw[textbar] (0.48,2.54) -- (1.22,2.54);
\draw[textbar] (0.48,2.28) -- (1.34,2.28);

\draw[flow] (1.74,2.79) -- (2.66,2.79);

\draw[doc] (2.82,1.98) rectangle (4.12,3.60);
\foreach \y in {3.30, 2.79, 2.28} {
    \draw[draw=black!28, fill=black!4, rounded corners=1.5pt, line width=.6pt]
        (2.98,\y-0.12) rectangle (3.96,\y+0.12);
    \draw[draw=black!45, line width=1.15pt, line cap=round]
        (3.10,\y) -- (3.84,\y);
}
\foreach \y in {3.045, 2.535} {
    \draw[badred, line width=1.1pt, line cap=round]
        (3.405,\y-0.065) -- (3.535,\y+0.065)
        (3.405,\y+0.065) -- (3.535,\y-0.065);
}

% Panel (a) metric pills (Zero-Shot: 0.87 / 1.5)
\node[pill, draw=black!25, fill=black!3, text=black!78] at (1.12,0.62) {
    {\scriptsize BERTScore}\\[-1pt]
    {\small\bfseries 0.87}
};
\node[pill, draw=badred!80, fill=badred!9, text=badred!90!black] at (3.28,0.62) {
    {\scriptsize Expert LPC}\\[-1pt]
    {\small\bfseries 1.5 / 5}
};

% ============================================================
% Panel (b): Concept graph G* -> ordered section subgraphs
% ============================================================
\node[cnode] (cF) at (5.22,3.44) {F};
\node[cnode] (cP) at (6.16,3.68) {P};
\node[cnode] (cS) at (7.10,3.24) {S};
\node[cnode] (cI) at (8.04,3.68) {I};
\node[cnode] (cE) at (8.98,3.24) {E};

\draw[->, >=stealth, draw=goodgreen, line width=.9pt] (cF) -- (cP);
\draw[->, >=stealth, draw=goodgreen, line width=.9pt] (cP) -- (cS);
\draw[->, >=stealth, draw=goodgreen, line width=.9pt] (cS) -- (cI);
\draw[->, >=stealth, draw=goodgreen, line width=.9pt] (cI) -- (cE);
\draw[->, >=stealth, draw=goodgreen!70, densely dashed, line width=.75pt]
    (cS) to[bend right=16] (cE);

\node[tinycap] at (5.22,2.98) {$G^*$};

\draw[flow] (7.10,2.94) -- (7.10,2.50);

% Ordered section boxes (internal edges + canonical order T)
\draw[draw=black!35, fill=black!3, rounded corners=3pt, line width=.75pt]
    (4.92,1.48) rectangle (6.92,2.42);
\node[snode] (sF) at (5.26,2.04) {F};
\node[snode] (sP) at (5.92,1.80) {P};
\node[snode] (sS) at (6.58,2.04) {S};
\draw[->, >=stealth, draw=goodgreen, line width=.8pt] (sF) -- (sP);
\draw[->, >=stealth, draw=goodgreen, line width=.8pt] (sP) -- (sS);

\draw[flow] (6.98,1.95) -- (7.48,1.95);
\node[tinycap] at (7.23,2.16) {$T$};

\draw[draw=black!35, fill=black!3, rounded corners=3pt, line width=.75pt]
    (7.54,1.48) rectangle (9.28,2.42);
\node[snode] (sI) at (7.98,2.04) {I};
\node[snode] (sE) at (8.84,1.80) {E};
\draw[->, >=stealth, draw=goodgreen, line width=.8pt] (sI) -- (sE);

% Panel (b) metric pills (FlowPlan-G2P: 0.83 / 4.7)
\node[pill, draw=black!25, fill=black!3, text=black!78] at (6.02,0.62) {
    {\scriptsize BERTScore}\\[-1pt]
    {\small\bfseries 0.83}
};
\node[pill, draw=goodgreen!85, fill=goodgreen!10, text=goodgreen!85!black] at (8.18,0.62) {
    {\scriptsize Expert LPC}\\[-1pt]
    {\small\bfseries 4.7 / 5}
};

\end{tikzpicture}%
}
\caption{Flat drafting versus graph-mediated planning. (a) Direct or chunked rewriting lacks a global plan. (b) \textsc{FlowPlan-G2P} partitions $G^*$ (F/P/S/I/E) into ordered section subgraphs ($T$).\looseness=-1}
\label{fig:intro_overview}

\endgroup
\end{figure}

Large language models (LLMs) have shown strong results on patent analytics tasks such as classification~\citep{emer2026large,yoo2025classification} and quality assessment~\citep{schmitt2025disentangling}. In automated patent drafting, however, prior work has heavily focused on claim generation and validation~\citep{Lee2020PatentClaim,Bai2024PatentGPT,li2024claimquality,jiang2025epd,yoo2025patentscore,yoo2026ace} as well as patent summarization~\citep{casola2022patentgen}, leaving long-form description generation far less explored. Systems that do target full descriptions generally assume patent-native inputs such as drafted claims and drawing text~\citep{patentformer2024}, or expand rough inventor disclosures along multi-agent or static guideline workflows~\citep{wang2024autopatent,jiang2026dis2pat}, leaving the paper-to-patent shift unaddressed. Directly targeting this shift, \textsc{PAP2PAT}~\citep{knappich2025pap2pat} handles long specifications through outline-guided chunk generation backed by local paper retrieval. Because each chunk is retrieved and realized locally without a global representation of the invention, cross-section technical dependencies break down, leading to repetitive phrasing, shallow implementation detail, or fine-tuning-induced hallucinations. As shown in Figure~\ref{fig:intro_overview}a, standard surface metrics such as ROUGE and BERTScore fail to detect these structural and legal defects, often favoring fluent but non-compliant drafts~\citep{jiang2025towards,pat_deval}.\looseness=-1

Preventing these structural and legal breakdowns requires modeling how invention components depend on one another globally before generating patent prose. We therefore propose \textsc{FlowPlan-G2P}, a graph-mediated framework that decouples technical reasoning from surface realization in three stages in Figure~\ref{fig:intro_overview}b. First, Concept Graph Induction extracts invention elements across canonical patent drafting categories and merges multi-candidate relational graphs into a directed concept graph $G^*$. Second, Section-Level Planning partitions $G^*$ into ordered section subgraphs using composite scoring over intra-section connectivity, category homogeneity, and explicit domain penalties, enforcing canonical section order and intra-section flow. Third, Graph-Conditioned Generation synthesizes patent descriptions from each linearized subgraph using section-specific drafting strategies and canonical formatting.

To evaluate generated descriptions beyond surface lexical overlap, we test \textsc{FlowPlan-G2P} on the 146-pair Pap2Pat-EvalGold benchmark~\citep{pat_deval}, alongside four-expert ratings on a subset of 75 paper--patent pairs. Comparing lexical scores against expert legal compliance reveals a sharp rank inversion where Zero-Shot leads in BERTScore at 0.8704 despite receiving an expert LPC score of 1.5/5 versus 4.7/5 for ours. Under expert-aligned Pat-DEVAL evaluation ($\tau \in [0.67, 0.76]$), \textsc{FlowPlan-G2P} outperforms all baselines across four dimensions, achieves an overall score of 4.3 with open-weight Llama-4-Scout compared to 2.3 for Few-Shot Sonnet-4.5, and exhibits complementary stage-wise gains in ablations.\looseness=-1

Our contributions are threefold.
\begin{itemize}
    \item \textbf{Graph-mediated formulation of paper-to-patent drafting.} Rather than drafting directly from raw paper text or retrieving chunks along a flat outline, \textsc{FlowPlan-G2P} introduces a two-level intermediate representation that induces a directed invention concept graph $G^*$ and partitions it into canonically ordered section subgraphs $\mathcal{P}$ before prose generation.\looseness=-1

    \item \textbf{Evidence that explicit planning narrows backbone gaps.} Single-pass Few-Shot prompting stays within 2.0--2.3 in overall Pat-DEVAL score across all tested LLMs, whereas adding \textsc{FlowPlan-G2P} lifts open-weight Llama-4-Scout to 4.3 (surpassing Few-Shot Sonnet-4.5 at 2.3) and scales further to 4.6 on DeepSeek-v3.1 and 4.7 on Sonnet-4.5.

    \item \textbf{Diagnostic insights on stage roles and metric--expert divergence.} Controlled ablations show that technical accuracy and legal compliance depend primarily on concept graph induction (Stage~1), whereas document layout depends most strongly on section-level planning (Stage~2). Moreover, while BERTScore and ROUGE-L over all 146 pairs rank Zero-Shot highest, four-expert evaluations on a subset of 75 paper--patent pairs assign Zero-Shot the lowest legal compliance.
\end{itemize}
\section{Background}
\label{sec:background}

Converting a scientific paper into a patent description requires reorganizing technical knowledge under statutory rules that govern whether an invention can be legally protected. Table~\ref{tab:paper_vs_patent} summarizes the key contrasts that motivate our structured framework.\looseness=-1

\begin{table}[h]
\centering
\small
\resizebox{\linewidth}{!}{
\begin{tabular}{lll}
\toprule
\textbf{Dimension} & \textbf{Scientific Paper} & \textbf{Patent Description} \\
\midrule
Rhetorical goal & Persuade peers of novelty & Disclose scope for protection \\
Information flow & Compress into key findings & Expand into full embodiments \\
Structure & Flexible narrative sections & Rigid canonical section order \\
Validity criterion & Empirical rigor & Statutory enablement \\
\bottomrule
\end{tabular}
}
\caption{Structural and rhetorical contrasts between scientific papers and patent descriptions.}
\label{tab:paper_vs_patent}
\end{table}

\paragraph{Rhetorical Shift and Expansion.}
Scientific papers argue for theoretical or empirical novelty, foregrounding benchmark gains and concise methodological summaries~\citep{murray2002innovation}. Patent descriptions operate as legal disclosure instruments that define utilitarian functions and concrete operational mechanisms~\citep{Althammer2021LinguisticallyInformedMasking}. Whereas an academic paper compresses implementation details to fit page budgets, a patent description must expand every technical component and its variants so that claim terms remain unambiguously supported.

\paragraph{Statutory Enablement.}
Under patent statutes such as 35 U.S.C.~\S~112 and Article~83 of the European Patent Convention, the description must disclose the invention clearly and completely enough for a Person Having Ordinary Skill in the Art (PHOSITA) to reproduce it without undue experimentation~\citep{jiang2024nlpinsurveys}. Direct rewriting or summarization with LLMs routinely violates this standard by omitting intermediate steps, parameter dependencies, or alternative embodiments needed for enablement.

\paragraph{Canonical Section Structure.}
A patent specification separates legal boundaries (claims) from technical exposition (the description), which serves as the technical dictionary and implementation guide supporting those claims. Standard descriptions follow a rigid macro-order covering \textit{Field}, \textit{Background}, \textit{Summary}, \textit{Detailed Description}, and \textit{Effects}, coupled with an intra-section progression from prior-art limitations to solutions, embodiments, and technical effects. End-to-end or locally chunked generation struggles to preserve these cross-section dependencies over long documents, motivating an explicit intermediate plan before text realization.
\section{Related Work}
\label{sec:related_work}

\subsection{Automated Patent Drafting and Paper-to-Patent Transfer}
Research on automated patent generation has focused primarily on individual specification components, especially patent claims and summaries. Early neural methods fine-tuned language models to generate claims from text inputs~\citep{Lee2020PatentClaim} and built large-scale summarization benchmarks~\citep{sharma2019bigpatent,casola2022patentgen}. Recent LLM-based studies extend claim drafting across jurisdictions and evaluate claim fidelity~\citep{Bai2024PatentGPT,li2024claimquality,jiang2025epd}, while multimodal systems target figure descriptions~\citep{shukla2025patentlmm}.

Full-specification generation presents greater structural demands. Patentformer drafts specifications from existing claims and drawing text~\citep{patentformer2024}, whereas multi-agent frameworks such as AutoPatent~\citep{wang2024autopatent} and Patent-MAF~\citep{jiang2026dis2pat} expand rough inventor disclosures through structured drafting and review loops. Although bibliometric studies have long documented knowledge transfer from scientific publications to patents~\citep{murray2002innovation,magerman2015involvement}, generative paper-to-patent transformation has received attention only recently. \textsc{PAP2PAT}~\citep{knappich2025pap2pat} introduces a large paper--patent corpus and handles long descriptions via outline-guided chunk generation with local paper retrieval, while RAG pipelines retrieve prior art to improve factual grounding~\citep{ding2025automatic,hindi2025enhancing}. However, local chunk retrieval does not maintain a global representation of cross-section technical dependencies, and RAG addresses external factual accuracy rather than internal section-level logical organization.

\subsection{Graph-Guided Planning for Structured Generation}
To improve coherence and factual fidelity in long-form generation, prior NLP work decouples content planning from surface realization. In data-to-text and graph-to-text settings, conditioning pretrained language models on explicit knowledge graphs or content plans reduces hallucination and preserves relational structure~\citep{puduppully2019data,zhao2020bridging,ribeiro2021investigating}. Plan-based frameworks such as Plan-and-Write~\citep{yao2019plan} and PAIR~\citep{hua2020pair} similarly organize long narratives through intermediate outlines before text generation.

In the patent domain, structured representations have recently improved analytical and decision-support tasks. PatentMind constructs multi-aspect reasoning graphs to evaluate patent similarity across technical features, domains, and claim scope~\citep{yoo2025patentmind}, while ERA aligns semantic patent representations with revealed economic preference for explainable valuation~\citep{yoo2025valuation}. Closer to generation, SPG couples pointer-based topology prediction with autoregressive decoding to enforce hierarchical dependency forests across patent claims~\citep{yoo2026spg}. Yet structure-aware planning for long-form patent descriptions from scientific papers remains underexplored, where intermediate plans must enforce canonical section order and legal enablement dependencies across prose sections rather than claim-level dependency trees. \textsc{FlowPlan-G2P} addresses this setting by combining patent-category concept graph induction with penalty-regularized section-level subgraph planning.

\subsection{Evaluation of Generated Patent Texts}
Standard generation metrics such as BLEU~\citep{papineni2002bleu}, ROUGE~\citep{lin2004rouge}, and BERTScore~\citep{zhang2020bertscore} measure lexical or embedding overlap with reference texts, which correlates poorly with human judgments when structural validity matters~\citep{reiter2018structured}. General-purpose LLM-as-a-judge frameworks, including G-Eval~\citep{liu2023geval} and LLM-Rubric~\citep{hashemi2024llmrubric}, improve alignment through rubric-guided scoring, but they do not encode patent-specific legal standards.

Recent work in patent NLP therefore emphasizes expert-grounded and multi-dimensional evaluation. For patent claims, \citet{li2024claimquality} combine expert reviews of feature coverage, conceptual clarity, and technical coherence to expose errors missed by n-gram metrics, while PatClaimEval~\citep{jiang2025towards} and PatentScore~\citep{yoo2025patentscore} introduce automated multi-dimensional evaluators aligned with patent experts. Broader patent drafting benchmarks similarly evaluate robustness, factual faithfulness, and structural compliance beyond lexical overlap~\citep{shomee2025patentwriter,knappich2025pap2pat,jiang2026dis2pat}. For long-form patent descriptions specifically, Pat-DEVAL~\citep{pat_deval} operationalizes PHOSITA-oriented Chain-of-Legal-Thought evaluation across technical fidelity, data precision, structural coverage, and legal compliance. Consistent with these studies, our experiments find that BERTScore and ROUGE-L favor Zero-Shot across all 146 pairs despite its lowest expert legal compliance rating on the 75-pair subset, motivating our use of Pat-DEVAL alongside human expert validation.
\section{Methodology}
\label{sec:methodology}

\begin{figure*}[t]
\centering
\begingroup

\definecolor{relsolves}{HTML}{55866A}
\definecolor{relimpl}{HTML}{C58B43}
\definecolor{relcauses}{HTML}{487FA8}
\definecolor{relimprov}{HTML}{8265A2}
\definecolor{relvalid}{HTML}{BF6262}

\resizebox{\textwidth}{!}{%
\begin{tikzpicture}[
    font=\normalsize,
    stagehead/.style={
        font=\bfseries\normalsize,
        text=black!88,
        align=center,
        inner sep=0pt
    },
    boxtitle/.style={
        font=\small\bfseries,
        text=black!85,
        align=center,
        inner sep=1pt
    },
    subtext/.style={
        font=\footnotesize,
        text=black!78,
        align=center,
        inner sep=1pt
    },
    tinynote/.style={
        font=\footnotesize,
        text=black!58,
        align=center,
        inner sep=1pt
    },
    panel/.style={
        draw=black!30,
        fill=black!2,
        rounded corners=5pt,
        line width=.85pt
    },
    card/.style={
        draw=black!45,
        fill=white,
        rounded corners=3.5pt,
        line width=.8pt
    },
    cardtint/.style={
        draw=black!55,
        fill=black!6,
        rounded corners=3.5pt,
        line width=.85pt
    },
    flow/.style={
        ->,
        >=stealth,
        draw=black!48,
        line width=.95pt,
        rounded corners=3pt
    },
    mainflow/.style={
        ->,
        >=stealth,
        draw=black!58,
        line width=1.25pt,
        rounded corners=3pt
    },
    gnode/.style={
        circle,
        draw=black!60,
        fill=white,
        minimum size=.54cm,
        inner sep=0pt,
        outer sep=0pt,
        line width=.85pt,
        font=\scriptsize\bfseries,
        text=black!85
    },
    gnodeaux/.style={
        circle,
        draw=black!42,
        fill=black!4,
        minimum size=.54cm,
        inner sep=0pt,
        outer sep=0pt,
        line width=.8pt,
        font=\scriptsize\bfseries,
        text=black!75
    },
    seccard/.style={
        draw=black!40,
        fill=white,
        rounded corners=3pt,
        line width=.75pt
    },
    pill/.style={
        font=\footnotesize,
        draw=black!32,
        fill=black!5,
        text=black!80,
        rounded corners=2pt,
        inner sep=2.5pt
    }
]

% ============================================================
% Canvas (24.30 x 8.20 -> crisp ~0.66x scale to \textwidth)
% ============================================================
\path[use as bounding box]
    (-.10,-1.05) rectangle (24.20,7.15);

% ============================================================
% Zone 0: Input Paper D
% ============================================================
\node[stagehead] at (1.15,6.82) {
    Input
};

% Folded document icon
\draw[draw=black!50, fill=white, line width=.95pt, line join=round]
    (.45,3.88) -- (.45,5.55) -- (1.52,5.55) -- (1.85,5.22) -- (1.85,3.88) -- cycle;
\draw[draw=black!50, line width=.95pt, line join=round]
    (1.52,5.55) -- (1.52,5.22) -- (1.85,5.22);

\foreach \y in {5.00, 4.72, 4.44, 4.16} {
    \draw[black!45, line width=1.2pt, line cap=round]
        (.68,\y) -- (1.60,\y);
}

\node[boxtitle] at (1.15,3.34) {
    Scientific\\Paper $D$
};

\draw[draw=black!25, fill=black!3, rounded corners=3pt, line width=.75pt]
    (.12,1.82) rectangle (2.18,2.74);
\node[subtext] at (1.15,2.28) {
    9 categories\\$\mathcal{S}_1, \dots, \mathcal{S}_9$
};

% Main arrow: Input -> Stage 1
\draw[mainflow] (2.28,3.45) -- (2.82,3.45);

% ============================================================
% Zone 1: (1) Concept Graph Induction (width 9.30 cm)
% ============================================================
\node[stagehead] at (7.50,6.82) {
    (1) Concept Graph Induction
};

\draw[panel] (2.85,.35) rectangle (12.15,6.45);

% Col 1a: Sequential reasoning (width 3.34 cm)
\draw[card] (3.08,4.55) rectangle (6.42,6.18);
\node[boxtitle] at (4.75,5.86) {
    Sequential LLM
};
\node[subtext] at (4.75,5.35) {
    $R_i = \mathrm{LLM}(D, R_{1:i-1})$
};
\node[tinynote] at (4.75,4.86) {
    Boilerplate-primed
};

\draw[flow] (4.75,4.55) -- (4.75,4.02);

% Col 1a: 3 candidate graphs
\draw[cardtint] (3.08,3.16) rectangle (6.42,3.98);
\node[boxtitle] at (4.75,3.73) {
    $G_1$ (Rule-based)
};
\node[tinynote] at (4.75,3.36) {
    7 canonical edge rules
};

\draw[card] (3.08,2.14) rectangle (6.42,2.96);
\node[boxtitle] at (4.75,2.71) {
    $G_2$ (LLM JSON)
};
\node[tinynote] at (4.75,2.34) {
    Implicit relations
};

\draw[card] (3.08,1.12) rectangle (6.42,1.94);
\node[boxtitle] at (4.75,1.69) {
    $G_3$ (LLM JSON)
};
\node[tinynote] at (4.75,1.32) {
    Implicit relations
};

\node[tinynote] at (4.75,.72) {
    3 candidate graphs
};

% Clean collector bus: G1, G2, G3 -> Deterministic Merge & Repair
\draw[draw=black!48, line width=.95pt]
    (6.42,3.57) -- (6.72,3.57)
    (6.42,2.55) -- (6.72,2.55)
    (6.42,1.53) -- (6.72,1.53)
    (6.72,1.53) -- (6.72,3.57);
\draw[flow]
    (6.72,2.55) -- (6.72,5.36) -- (7.10,5.36);

% Col 1b: Deterministic Merge & Repair (width 4.80 cm)
\draw[card] (7.12,4.55) rectangle (11.92,6.18);
\node[boxtitle] at (9.52,5.86) {
    Deterministic Merge \& Repair
};
\node[subtext] at (9.52,5.35) {
    Node union + edge majority vote
};
\node[tinynote] at (9.52,4.86) {
    Mandatory check + isolated repair
};

\draw[flow] (9.52,4.55) -- (9.52,4.02);

% Col 1b: Complete 9-node Concept Graph G*
\draw[draw=black!45, fill=white, densely dashed, rounded corners=4pt, line width=.85pt]
    (7.12,.62) rectangle (11.92,3.98);
\node[boxtitle] at (9.52,3.68) {
    Concept Graph $G^* = (V^*, E^*)$
};

% All 9 canonical nodes
\node[gnodeaux] (nF)   at (7.62,3.04) {F};
\node[gnodeaux] (nPA)  at (7.62,2.02) {PA};
\node[gnode]    (nNov) at (9.02,3.08) {Nov};
\node[gnode]    (nTP)  at (8.74,2.08) {TP};
\node[gnode]    (nSol) at (10.02,2.34) {Sol};
\node[gnode]    (nImp) at (9.58,1.16) {Imp};
\node[gnode]    (nEmb) at (10.88,1.08) {Emb};
\node[gnode]    (nFig) at (11.38,1.94) {Fig};
\node[gnode]    (nEff) at (11.30,3.04) {Eff};

% 7 exact rule-based directed edges (color reserved for relation semantics)
\draw[->, >=stealth, draw=relsolves, line width=1.05pt] (nTP)  -- (nSol);
\draw[->, >=stealth, draw=relimpl,   line width=1.05pt] (nSol) -- (nImp);
\draw[->, >=stealth, draw=relcauses, line width=1.05pt] (nImp) -- (nEff);
\draw[->, >=stealth, draw=relimprov, densely dashed, line width=1.0pt] (nSol) -- (nEff);
\draw[->, >=stealth, draw=relimprov, densely dashed, line width=1.0pt] (nNov) -- (nSol);
\draw[->, >=stealth, draw=relvalid,  line width=1.05pt] (nEmb) -- (nImp);
\draw[->, >=stealth, draw=relvalid,  line width=1.05pt] (nFig) -- (nEmb);

% Main arrow: Stage 1 -> Stage 2
\draw[mainflow] (12.15,3.45) -- (12.67,3.45);

% ============================================================
% Zone 2: (2) Section-Level Planning (width 5.50 cm)
% ============================================================
\node[stagehead] at (15.45,6.82) {
    (2) Section-Level Planning
};

\draw[panel] (12.70,.35) rectangle (18.20,6.45);

% Candidate Sampling + Gating + Penalty Box
\draw[card] (12.92,4.78) rectangle (17.98,6.18);
\node[boxtitle] at (15.45,5.88) {
    $k = 5$ Candidate Plans \& Scoring
};
\node[subtext] at (15.45,5.44) {
    $C \ge 0.5,\;\; Sim \ge 0.6$
};
\node[subtext] at (15.45,5.04) {
    $\mathrm{Score}(\mathcal{P}) = C + Sim - P_{\mathrm{pen}}$
};

\draw[flow] (15.45,4.78) -- (15.45,4.38);

% 5 Canonical Ordered Section Subgraphs
\draw[seccard] (12.92,3.80) rectangle (17.98,4.36);
\node[anchor=west, font=\footnotesize\bfseries, text=black!85] at (13.10,4.08) {1. Field};
\node[anchor=east, pill] at (17.82,4.08) {F};

\draw[seccard] (12.92,3.08) rectangle (17.98,3.64);
\node[anchor=west, font=\footnotesize\bfseries, text=black!85] at (13.10,3.36) {2. Background};
\node[anchor=east, pill] at (17.82,3.36) {PA, TP};

\draw[seccard] (12.92,2.36) rectangle (17.98,2.92);
\node[anchor=west, font=\footnotesize\bfseries, text=black!85] at (13.10,2.64) {3. Summary};
\node[anchor=east, pill] at (17.82,2.64) {Sol, Nov};

\draw[seccard] (12.92,1.64) rectangle (17.98,2.20);
\node[anchor=west, font=\footnotesize\bfseries, text=black!85] at (13.10,1.92) {4. Detailed Desc.};
\node[anchor=east, pill] at (17.82,1.92) {Imp, Emb, Fig};

\draw[seccard] (12.92,.92) rectangle (17.98,1.48);
\node[anchor=west, font=\footnotesize\bfseries, text=black!85] at (13.10,1.20) {5. Effects};
\node[anchor=east, pill] at (17.82,1.20) {Eff};

% Canonical order connectors
\foreach \ytop/\ybot in {3.80/3.64, 3.08/2.92, 2.36/2.20, 1.64/1.48} {
    \draw[->, >=stealth, draw=black!42, line width=.75pt]
        (15.45,\ytop) -- (15.45,\ybot);
}

\node[tinynote] at (15.45,.62) {
    Ordered subgraphs $S_1, \dots, S_5$
};

% Main arrow: Stage 2 -> Stage 3
\draw[mainflow] (18.20,3.45) -- (18.72,3.45);

% ============================================================
% Zone 3: (3) Graph-Conditioned Generation (width 5.30 cm)
% ============================================================
\node[stagehead] at (21.40,6.82) {
    (3) Graph-Conditioned Generation
};

\draw[panel] (18.75,.35) rectangle (24.05,6.45);

% Step 3a: Linearization & Prompting
\draw[card] (18.97,4.72) rectangle (23.83,6.18);
\node[boxtitle] at (21.40,5.88) {
    Linearization \& Prompting
};
\node[subtext] at (21.40,5.42) {
    $\mathrm{Linearize}(S_i)$ (Concepts, Relations)
};
\node[tinynote] at (21.40,5.00) {
    PatentInstruction + $\mathrm{Strategy}_i$
};

\draw[flow] (21.40,4.72) -- (21.40,4.28);

% Step 3b: LLM Synthesis
\draw[cardtint] (19.25,3.54) rectangle (23.55,4.24);
\node[boxtitle] at (21.40,3.89) {
    LLM Synthesis ($T_{\mathrm{gen}} = 0.2$)
};

\draw[flow] (21.40,3.54) -- (21.40,3.12);

% Step 3c: Final Assembled Patent Description
\draw[draw=black!58, fill=white, rounded corners=3.5pt, line width=.95pt]
    (18.97,.62) rectangle (23.83,3.08);

\node[boxtitle] at (21.40,2.82) {
    PATENT DESCRIPTION
};
\draw[black!18, line width=.65pt] (19.20,2.60) -- (23.60,2.60);

\node[anchor=west, font=\footnotesize\bfseries, text=black!80] at (19.16,2.32) {1. FIELD};
\draw[black!38, line width=1.2pt, line cap=round] (22.30,2.32) -- (23.62,2.32);

\node[anchor=west, font=\footnotesize\bfseries, text=black!80] at (19.16,1.94) {2. BACKGROUND};
\draw[black!38, line width=1.2pt, line cap=round] (22.30,1.94) -- (23.62,1.94);

\node[anchor=west, font=\footnotesize\bfseries, text=black!80] at (19.16,1.56) {3. SUMMARY};
\draw[black!38, line width=1.2pt, line cap=round] (22.30,1.56) -- (23.62,1.56);

\node[anchor=west, font=\footnotesize\bfseries, text=black!80] at (19.16,1.18) {4. DETAILED DESC.};
\draw[black!38, line width=1.2pt, line cap=round] (22.30,1.18) -- (23.62,1.18);

\node[anchor=west, font=\footnotesize\bfseries, text=black!80] at (19.16,.80) {5. EFFECTS};
\draw[black!38, line width=1.2pt, line cap=round] (22.30,.80) -- (23.62,.80);

% ============================================================
% Dedicated bottom legend strip (2 clean centered rows)
% ============================================================
\draw[black!14, line width=.65pt] (.15,.04) -- (24.00,.04);

% Row 1: 5 exact relation types
\node[anchor=west, font=\footnotesize\bfseries, text=black!75] at (2.40,-.30) {
    Edge relations in $G^*$
};

\draw[->, >=stealth, draw=relsolves, line width=1.15pt] (6.40,-.30) -- (7.10,-.30);
\node[anchor=west, font=\footnotesize, text=black!82] at (7.22,-.30) {\textit{solves}};

\draw[->, >=stealth, draw=relimpl, line width=1.15pt] (9.20,-.30) -- (9.90,-.30);
\node[anchor=west, font=\footnotesize, text=black!82] at (10.02,-.30) {\textit{implements}};

\draw[->, >=stealth, draw=relcauses, line width=1.15pt] (12.60,-.30) -- (13.30,-.30);
\node[anchor=west, font=\footnotesize, text=black!82] at (13.42,-.30) {\textit{causes}};

\draw[->, >=stealth, draw=relimprov, densely dashed, line width=1.1pt] (15.40,-.30) -- (16.10,-.30);
\node[anchor=west, font=\footnotesize, text=black!82] at (16.22,-.30) {\textit{improves}};

\draw[->, >=stealth, draw=relvalid, line width=1.15pt] (18.60,-.30) -- (19.30,-.30);
\node[anchor=west, font=\footnotesize, text=black!82] at (19.42,-.30) {\textit{validates}};

% Row 2: All 9 node category abbreviations
\node[font=\footnotesize, text=black!65, align=center] at (12.05,-.76) {
    \textbf{Categories}\quad F Field \textbullet\ PA PriorArt \textbullet\ TP TechProblem \textbullet\ Nov Novelty \textbullet\ Sol Solution \textbullet\ Imp Implementation \textbullet\ Emb Embodiment \textbullet\ Fig Figure \textbullet\ Eff Effects
};

\end{tikzpicture}%
}

\caption{Overview of \textsc{FlowPlan-G2P}. Stage~1 induces and deterministically merges a nine-category concept graph $G^*$ from a scientific paper $D$. Stage~2 samples candidate plans scored by connectivity, homogeneity, and domain penalties to produce five ordered section subgraphs. Stage~3 linearizes each subgraph and generates canonical patent sections under section-specific drafting strategies.\looseness=-1}
\label{fig:flowplan_g2p_architecture}

\endgroup
\end{figure*}

\textsc{FlowPlan-G2P} transforms a scientific paper into a structured patent description through three stages shown in Figure~\ref{fig:flowplan_g2p_architecture}. Stage~1 induces a directed concept graph over patent drafting categories as detailed in Section~\ref{sec:stage1}. Stage~2 partitions that graph into ordered section subgraphs via penalty-regularized scoring as described in Section~\ref{sec:stage2}, and Stage~3 generates section-specific patent text conditioned on each linearized subgraph as presented in Section~\ref{sec:stage3}.

\subsection{Concept Graph Induction}
\label{sec:stage1}

Given an input paper $D$, we first decompose its technical content sequentially across nine canonical patent drafting categories
\begin{align}
\mathcal{S} = \{ & \textit{Field}, \textit{TechProblem}, \textit{PriorArt}, \notag \\
& \textit{Novelty}, \textit{Solution}, \textit{Implementation}, \notag \\
& \textit{Effects}, \textit{Embodiment}, \textit{Figure} \}.
\end{align}
For each category $S_i \in \mathcal{S}$, the LLM generates a structured reasoning output $R_i$ conditioned on $D$ and truncated previews (up to 200 characters per category) of prior outputs $R_{1:i-1}$,
\begin{equation}
R_i = \mathrm{LLM}(\mathrm{prompt}_i(D, R_{1:i-1})).
\end{equation}
Each prompt primes patent register with a category-specific opening boilerplate, such as ``The present invention relates to'' for \textit{Field} and ``However, conventional technology has the following drawbacks'' for \textit{TechProblem} in Appendix~\ref{app:expert_prompts}. Conditioning on prior steps maintains causal continuity as the model moves from prior-art limitations to solutions, embodiments, and effects.

From $\{R_i\}_{i=1}^{9}$, we build a directed concept graph $G = (V, E)$, where each node $v \in V$ stores a category label $\tau(v) \in \mathcal{S}$ and its extracted content, and each directed edge $(u, v, r) \in E$ carries a functional relation $r \in \mathcal{R}$,
\begin{equation}
\begin{aligned}
\mathcal{R} = \{&
  \textit{solves}, \textit{implements}, \textit{causes},\\
& \textit{improves}, \textit{validates}\}.
\end{aligned}
\end{equation}

Directed edges follow the forward patent drafting progression, such as $\textit{TechProblem} \xrightarrow{\textit{solves}} \textit{Solution}$ and $\textit{Solution} \xrightarrow{\textit{implements}} \textit{Implementation}$.

To balance structural precision with recall of implicit dependencies, we generate three candidate graphs $\{G_1, G_2, G_3\}$. Candidate $G_1$ is constructed deterministically by creating one node per non-empty category and applying seven canonical edge templates ($\textit{TechProblem} \xrightarrow{\textit{solves}} \textit{Solution}$, $\textit{Solution} \xrightarrow{\textit{implements}} \textit{Implementation}$, $\textit{Implementation} \xrightarrow{\textit{causes}} \textit{Effects}$, $\textit{Solution} \xrightarrow{\textit{improves}} \textit{Effects}$, $\textit{Embodiment} \xrightarrow{\textit{validates}} \textit{Implementation}$, $\textit{Figure} \xrightarrow{\textit{validates}} \textit{Embodiment}$, and $\textit{Novelty} \xrightarrow{\textit{improves}} \textit{Solution}$). Candidates $G_2$ and $G_3$ prompt the LLM to output structured JSON graphs over 300-character summaries of $\{R_i\}$ to capture non-template relations, falling back to the rule-based graph on parsing failure.

We consolidate $\{G_1, G_2, G_3\}$ deterministically into $G^* = (V^*, E^*)$. Nodes are aggregated by identifier union across candidates, and conflicting relation labels for any ordered pair $(u, v)$ are resolved by majority voting. A post-merge repair step inserts placeholder nodes for any missing mandatory category in $\{\textit{Field}, \textit{TechProblem}, \textit{Solution}\}$ and links any isolated \textit{Effects} node from a \textit{Solution} node via a \textit{causes} edge.

\subsection{Paragraph and Section Planning}
\label{sec:stage2}

Stage~2 partitions $G^*$ into an ordered plan $\mathcal{P} = (\{S_1, \dots, S_M\}, T)$ aligned with five canonical patent sections and their default category mappings, namely \textit{Field} ($\{\textit{Field}\}$), \textit{Background} ($\{\textit{TechProblem}, \textit{PriorArt}\}$), \textit{Summary} ($\{\textit{Solution}, \textit{Novelty}\}$), \textit{Detailed Description} ($\{\textit{Implementation}, \textit{Embodiment}, \textit{Figure}\}$), and \textit{Effects} ($\{\textit{Effects}\}$). To generate a candidate plan, we prompt the LLM with node previews (150 characters per node) and all directed edges of $G^*$ to assign each node to one canonical section. Unassigned nodes or malformed JSON outputs fall back deterministically to the default category-to-section mapping above.

We sample $k = 5$ candidate plans and evaluate each candidate over multi-node sections $\mathcal{M}_{\ge 2} = \{i \mid |V(S_i)| \ge 2\}$. Intra-section connectivity $C$ averages directed edge density across $\mathcal{M}_{\ge 2}$,
\begin{equation}
\begin{aligned}
C_i &= \frac{|E_{\text{in}}(S_i)|}
{|V(S_i)|\bigl(|V(S_i)| - 1\bigr)}, \\
C &= \frac{1}{\max(1, |\mathcal{M}_{\ge 2}|)}
\sum_{i \in \mathcal{M}_{\ge 2}} C_i,
\end{aligned}
\end{equation}
where $E_{\text{in}}(S_i)$ denotes directed edges whose
endpoints both belong to $S_i$.
Category homogeneity $Sim$ averages normalized Shannon
entropy across $\mathcal{M}_{\ge 2}$,
\begin{equation}
\begin{aligned}
Sim_i &= 1 - \frac{H(S_i)}
{\log_2 |\mathcal{T}(S_i)|}, \\
Sim &= \frac{1}{\max(1, |\mathcal{M}_{\ge 2}|)}
\sum_{i \in \mathcal{M}_{\ge 2}} Sim_i,
\end{aligned}
\end{equation}

where $H(S_i)$ is the empirical entropy of node category counts over the set of distinct categories $\mathcal{T}(S_i)$ in $S_i$, with $Sim_i = 1$ when $|\mathcal{T}(S_i)| = 1$.

To enforce drafting conventions, we compute an explicit domain penalty $P_{\text{pen}}$ that adds $1.0$ for each missing mandatory category in $\{\textit{Field}, \textit{TechProblem}, \textit{Solution}\}$, $0.5$ for each prohibited co-occurrence of $(\textit{Field}, \textit{Effects})$ or $(\textit{PriorArt}, \textit{Solution})$ within the same section, and $0.3$ for each canonical section-order inversion. Each candidate receives combined score

\begin{equation}
\mathrm{Score}(\mathcal{P}) = C + Sim - P_{\text{pen}}.
\end{equation}

Candidates passing both thresholds $C \ge \tau_C$ and $Sim \ge \tau_S$ are ranked by $\mathrm{Score}(\mathcal{P})$; if no candidate meets both thresholds, the highest-scoring candidate is selected as fallback. We set $\tau_C = 0.5$ and $\tau_S = 0.6$ via grid search over $\tau_C \in \{0.3, \dots, 0.7\}$ and $\tau_S \in \{0.4, \dots, 0.8\}$ on a 20-pair development set, where LPC varies by less than $0.3$ across $\tau_C \in [0.4, 0.6]$ and $\tau_S \in [0.5, 0.7]$. Note that $C_i$ counts intra-section edges only; cross-section links such as $\textit{TechProblem} \xrightarrow{\textit{solves}} \textit{Solution}$ remain in $G^*$ and in the Stage~2 planning input, whereas Stage~3 receives each section's internal edges $E_i$ arranged in canonical section order $T$. Moreover, because $Sim_i$ evaluates to $0$ when a multi-category section holds one node per category (as in a two-node \textit{Background} or \textit{Summary}), balanced canonical candidates can fail to meet both thresholds; whenever no candidate passes both gates, fallback selection picks the highest-scoring candidate under $\mathrm{Score}(\mathcal{P})$, where $P_{\text{pen}}$ and canonical ordering govern the final layout. Finally, empty sections are dropped, induced subgraphs are extracted from $G^*$, and the surviving sections are sorted in canonical order.\looseness=-1

\subsection{Graph-Conditioned Generation}
\label{sec:stage3}

For each ordered section subgraph $S_i = (V_i, E_i)$, we serialize its structure into a structured text block $\mathrm{Linearize}(S_i)$ consisting of a concept list (each node formatted with its identifier, category, and content) followed by a relation list of all internal directed edges $(u, r, v) \in E_i$. Each section is synthesized by concatenating a shared patent drafting template, a section-specific strategy block, and the linearized subgraph,
\begin{equation}
\begin{aligned}
\mathrm{Prompt}(S_i) = [&\mathrm{PatentInstruction};\\
&\mathrm{Strategy}_i;\, \mathrm{Linearize}(S_i)].
\end{aligned}
\end{equation}

The shared template enforces formal patent register, explicit opening boilerplates, repetitive elaboration across alternative embodiments, and figure references (Appendix~\ref{appendix:prompt}). Each strategy $\mathrm{Strategy}_i$ appends section-specific requirements aligned with legal function. \textit{Field} emphasizes concise technical scope, \textit{Background} frames prior-art limitations without prematurely disclosing the invention, \textit{Summary} highlights core technical mechanisms and objects of the invention, \textit{Detailed Description} expands multiple embodiments with explicit figure references (such as ``Figure~1 illustrates...''), and \textit{Effects} summarizes concrete technical advantages. Sections are generated sequentially with low-temperature decoding ($T_{\text{gen}} = 0.2$ where supported, with up to three exponential-backoff API retries) and post-processed to standardize canonical uppercase section headers under a unified patent description.

\section{Dataset}
\label{sec:dataset}

We evaluate on Pap2Pat-EvalGold~\citep{pat_deval}, a high-precision subset curated from the \textsc{PAP2PAT} corpus~\citep{knappich2025pap2pat}. While \textsc{PAP2PAT} aligns scientific papers and patents at scale through automated matching, its full release retains pairs with loose technical or authorship correspondence. Pap2Pat-EvalGold filters the corpus to retain only pairs satisfying both strong semantic alignment (Sentence-BERT cosine similarity $\ge 0.8$~\citep{reimers2019sentencebert}) and verified authorship consistency (author overlap ratio $\ge 0.5$ between paper authors and patent inventors), yielding 146 paper--patent pairs.

Across our experiments, we use this curated benchmark together with four scoped evaluation subsets. All main automatic comparisons run on the full 146-pair set. Human evaluation covers a subset of 75 paper--patent pairs (roughly half the benchmark) independently rated by four patent experts. Component ablations use a 30-pair domain-stratified subset, and Stage~2 threshold selection ($\tau_C, \tau_S$) uses a separate 20-pair development set. To test sensitivity to curation, we also evaluate on 30 unfiltered pairs drawn directly from \textsc{PAP2PAT}.
\section{Experiments and Results}
\label{sec:experiments}

We empirically evaluate \textsc{FlowPlan-G2P} across metric alignment, baseline comparisons, backbone generalization, and component ablations.

\subsection{Evaluation Protocol and Metric Validation}
\label{sec:metric_validation}

Standard NLG metrics such as ROUGE and BERTScore measure surface lexical or embedding overlap, which cannot distinguish statutory patent descriptions from fluent paraphrases that omit enablement details. We therefore adopt Pat-DEVAL~\citep{pat_deval}, which uses Chain-of-Legal-Thought (CoLT) reasoning to approximate a PHOSITA review along four dimensions, namely Technical Content Fidelity (TCF), Data Precision (DP), Structural Coverage (SC), and Legal-Professional Compliance (LPC), each scored on a 1--5 scale.

\paragraph{Human Evaluation Protocol.}
Four patent experts independently evaluated generated descriptions on a subset of 75 paper--patent pairs (approximately 50\% of the benchmark). Each annotator scored every output along the same four Pat-DEVAL dimensions (TCF, DP, SC, and LPC) on a 5-point Likert scale, and final human scores average across all four annotators.

\paragraph{Inadequacy of Surface Metrics.}
Table~\ref{tab:metric_divergence} contrasts standard NLG scores on the full 146-pair set against expert Legal-Professional Compliance (Human-LPC) on the subset of 75 paper--patent pairs. Zero-Shot achieves the highest BERTScore (0.8704) and ROUGE-L (0.1780) despite receiving the lowest expert legal compliance rating (Human-LPC 1.5). Conversely, \textsc{FlowPlan-G2P} records lower BERTScore (0.8302) and ROUGE-L (0.1689) while reaching 4.7 in Human-LPC. Although ROUGE-1 and ROUGE-2 rank \textsc{FlowPlan-G2P} highest, the rank inversion under BERTScore and ROUGE-L shows that surface similarity does not track expert legal compliance ratings.

\begin{table}[h]
\centering
\footnotesize
\setlength{\tabcolsep}{4pt}
\begin{adjustbox}{max width=\columnwidth}
\begin{tabular}{lccccc}
\toprule
\textbf{Model} & \textbf{R-1} & \textbf{R-2} & \textbf{R-L} & \textbf{BS} & \textbf{H-LPC} \\
\midrule
Zero-Shot & 0.3591 & 0.1903 & \textbf{0.1780} & \textbf{0.8704} & 1.5 \\
Few-Shot & 0.3312 & 0.1224 & 0.1377 & 0.8337 & 2.1 \\
\textbf{\textsc{FlowPlan-G2P}} & \textbf{0.5446} & \textbf{0.2204} & 0.1689 & 0.8302 & \textbf{4.7} \\
\bottomrule
\end{tabular}
\end{adjustbox}
\caption{Surface metrics (ROUGE and BERTScore on all 146 pairs) versus expert legal compliance (H-LPC on a subset of 75 paper--patent pairs). Zero-Shot ranks highest in BERTScore and ROUGE-L despite the lowest expert H-LPC rating.}
\label{tab:metric_divergence}
\end{table}

\paragraph{Alignment of Pat-DEVAL with Experts.}
Table~\ref{tab:expert_correlation} reports expert ratings across all four dimensions alongside Kendall's $\tau$ correlation between automated Pat-DEVAL scores and human judgments. Correlations remain consistently strong across the full quality range, from $\tau = 0.67$ for Few-Shot and $\tau = 0.72$ for Zero-Shot up to $\tau = 0.76$ for \textsc{FlowPlan-G2P}. Based on this agreement, we use Pat-DEVAL as the primary automatic metric in subsequent evaluations.

\begin{table}[h]
\centering
\footnotesize
\setlength{\tabcolsep}{4.5pt}
\begin{adjustbox}{max width=\columnwidth}
\begin{tabular}{lccccc}
\toprule
& \multicolumn{4}{c}{\textbf{Human Expert Scores (1--5)}} & \\
\cmidrule(lr){2-5}
\textbf{Model} & \textbf{TCF} & \textbf{DP} & \textbf{SC} & \textbf{LPC} & \textbf{Kendall's $\tau$} \\
\midrule
Zero-Shot & 1.7 & 1.4 & 1.8 & 1.5 & 0.72 \\
Few-Shot & 2.3 & 2.0 & 2.4 & 2.1 & 0.67 \\
\textsc{PAP2PAT} & 3.4 & 3.1 & 3.3 & 3.0 & 0.69 \\
\textbf{\textsc{FlowPlan-G2P}} & \textbf{4.5} & \textbf{4.4} & \textbf{4.6} & \textbf{4.7} & \textbf{0.76} \\
\bottomrule
\end{tabular}
\end{adjustbox}
\caption{Human expert ratings and Pat-DEVAL rank correlation (Kendall's $\tau$) on a subset of 75 paper--patent pairs.}
\label{tab:expert_correlation}
\end{table}

\subsection{Baseline Comparison}
\label{sec:baseline_comparison}

We compare \textsc{FlowPlan-G2P} against Zero-Shot prompting, Few-Shot prompting, and \textsc{PAP2PAT} on all 146 pairs using Sonnet-4.5 as a shared backbone in Table~\ref{tab:baseline_comparison}. Direct prompting baselines score at or below 2.5 across every dimension; although they produce fluent prose, they routinely omit canonical patent sections, embodiment expansion, and explicit claim support, yielding weak Structural Coverage (SC) and Legal-Professional Compliance (LPC).

\begin{table}[h]
\centering
\footnotesize
\setlength{\tabcolsep}{6pt}
\begin{adjustbox}{max width=\columnwidth}
\begin{tabular}{lcccc}
\toprule
\textbf{Model} & \textbf{TCF} & \textbf{DP} & \textbf{SC} & \textbf{LPC} \\
\midrule
Zero-Shot Prompting & 1.8 & 1.5 & 1.9 & 1.6 \\
Few-Shot Prompting & 2.4 & 2.1 & 2.5 & 2.2 \\
\textsc{PAP2PAT} & 3.5 & 3.2 & 3.4 & 3.1 \\
\textbf{\textsc{FlowPlan-G2P} (Ours)} & \textbf{4.6} & \textbf{4.5} & \textbf{4.7} & \textbf{4.8} \\
\bottomrule
\end{tabular}
\end{adjustbox}
\caption{Pat-DEVAL scores on all 146 pairs using Sonnet-4.5 as the shared backbone. On 30 unfiltered \textsc{PAP2PAT} pairs, \textsc{FlowPlan-G2P} drops by 0.4 points on average and remains above 4.0.}
\label{tab:baseline_comparison}
\end{table}

\textsc{PAP2PAT} improves substantially over direct prompting (3.1--3.5) via outline-guided generation, yet still loses cross-section specificity and data precision. \textsc{FlowPlan-G2P} reaches at least 4.5 on all four metrics, scoring 4.6 in TCF, 4.5 in DP, 4.7 in SC, and 4.8 in LPC. Paired bootstrap resampling (1,000 resamples) confirms that improvements over \textsc{PAP2PAT} are statistically significant across all dimensions, with 95\% confidence intervals of $[+0.8, +1.4]$ for TCF, $[+1.0, +1.6]$ for DP, $[+1.0, +1.6]$ for SC, and $[+1.4, +2.0]$ for LPC. On 30 unfiltered pairs from \textsc{PAP2PAT}, average performance decreases by 0.4 points while staying above 4.0, indicating robustness outside the curated benchmark.

\subsection{Cross-Backbone Robustness}
\label{sec:llm_dependency}

To test whether gains depend on a proprietary backbone, Table~\ref{tab:llm_dependency} compares overall Pat-DEVAL scores (averaged over the four dimensions) across Llama-4-Scout, DeepSeek-v3.1, and Sonnet-4.5 under vanilla Few-Shot prompting versus \textsc{FlowPlan-G2P}.

\begin{table}[h]
\centering
\footnotesize
\setlength{\tabcolsep}{6pt}
\begin{adjustbox}{max width=\columnwidth}
\begin{tabular}{lcc}
\toprule
\textbf{Backbone} & \textbf{Vanilla (Few-Shot)} & \textbf{+ \textsc{FlowPlan-G2P}} \\
\midrule
Llama-4-Scout & 2.0 & 4.3 \\
DeepSeek-v3.1 & 2.2 & 4.6 \\
Sonnet-4.5 & 2.3 & 4.7 \\
\bottomrule
\end{tabular}
\end{adjustbox}
\caption{Overall Pat-DEVAL scores across open-weight and proprietary backbones on all 146 pairs.}
\label{tab:llm_dependency}
\end{table}

Two patterns emerge. First, open-weight Llama-4-Scout with \textsc{FlowPlan-G2P} reaches an overall Pat-DEVAL score of 4.3, outperforming Few-Shot Sonnet-4.5 at 2.3. Second, gains persist on stronger backbones, lifting DeepSeek-v3.1 from 2.2 to 4.6 and Sonnet-4.5 from 2.3 to 4.7.

\subsection{Ablation Study}
\label{sec:ablation}

To isolate stage-wise contributions, we evaluate three ablated variants on the 30-pair domain-stratified subset in Table~\ref{tab:ablation}; minor differences from Table~\ref{tab:baseline_comparison} reflect subset sampling variance. Specifically, we test (1) removing Concept Graph Induction (Stage~1) by generating sections directly from the paper with section prompts, (2) removing Section-Level Planning (Stage~2) by skipping section clustering over the merged graph, and (3) disabling post-merge verification and post-processing.

\begin{table}[h]
\centering
\footnotesize
\setlength{\tabcolsep}{4.5pt}
\begin{adjustbox}{max width=\columnwidth}
\begin{tabular}{lcccc}
\toprule
\textbf{Configuration} & \textbf{TCF} & \textbf{DP} & \textbf{SC} & \textbf{LPC} \\
\midrule
Full \textsc{FlowPlan-G2P} & \textbf{4.8} & \textbf{4.3} & \textbf{4.6} & \textbf{4.7} \\
w/o Concept Graph (Stage~1) & 3.1 & 2.8 & 3.3 & 3.2 \\
w/o Section Planning (Stage~2) & 3.8 & 3.5 & 3.2 & 3.6 \\
w/o Post-Processing \& Repair & 4.4 & 4.1 & 4.5 & 4.5 \\
\bottomrule
\end{tabular}
\end{adjustbox}
\caption{Ablation results on the 30-pair domain-stratified subset. Removing Stage~1 causes the largest overall drop, whereas removing Stage~2 degrades Structural Coverage most severely.}
\label{tab:ablation}
\end{table}

Removing Stage~1 causes the largest drop on TCF ($-1.7$), DP ($-1.5$), and LPC ($-1.5$) while reducing SC by $1.3$ points, lowering scores to 2.8--3.3 and confirming that the induced concept graph rather than section prompting alone drives technical fidelity. Removing Stage~2 produces the largest decline in Structural Coverage ($-1.4$ points, from 4.6 to 3.2), reflecting its role in enforcing canonical section layout and intra-section progression. Disabling verification and post-processing leads to smaller but consistent drops of 0.1--0.4 points across all dimensions, with the largest effect in TCF ($-0.4$).
\section{Discussion}
\label{sec:discussion}

\paragraph{Cross-backbone gains from structured planning.}
Across the three backbones in Table~\ref{tab:llm_dependency}, single-pass Few-Shot prompting remains within a narrow 2.0--2.3 band in overall Pat-DEVAL score, whereas \textsc{FlowPlan-G2P} raises scores to 4.3--4.7. Most of the gap in this comparison comes from adding explicit graph construction and section planning, while stronger backbones achieve higher overall scores within the same pipeline.

\paragraph{Distinct roles of Stage~1 and Stage~2.}
In Table~\ref{tab:ablation}, removing Stage~1 produces the largest declines in TCF ($-1.7$), DP ($-1.5$), and LPC ($-1.5$), whereas removing Stage~2 produces the largest decline in SC ($-1.4$). Concept extraction and graph consolidation primarily support technical fidelity, data precision, and legal compliance, while section partitioning, penalty-regularized selection, and canonical ordering govern structural coverage.

\paragraph{Divergence between surface metrics and expert ratings.}
Paper-to-patent transformation requires reorganizing concise research text into boilerplate-framed embodiments across standard patent sections. On Pap2Pat-EvalGold, BERTScore and ROUGE-L favored Zero-Shot (0.8704 BERTScore), which received the lowest expert LPC rating (1.5) on the evaluated subset, whereas \textsc{FlowPlan-G2P} scored lower in BERTScore (0.8302) and highest in expert LPC (4.7). Relying solely on lexical overlap metrics can therefore misalign system rankings with expert drafting evaluations.

\paragraph{Inspectable intermediate artifacts.}
Decomposing generation into three stages increases inference cost relative to single-pass prompting, but produces two structured artifacts prior to text synthesis. The intermediate concept graph $G^*$ and ordered section plan $\mathcal{P}$ expose extracted technical links and section assignments for inspection before drafting, allowing practitioners to spot missing components or misplaced sections without editing full-length prose.
\section{Conclusion}
\label{sec:conclusion}

We introduced \textsc{FlowPlan-G2P}, a graph-mediated framework that transforms scientific papers into structured patent descriptions through concept graph induction, section-level planning, and graph-conditioned generation. On Pap2Pat-EvalGold, BERTScore and ROUGE-L over all 146 pairs rank Zero-Shot highest, whereas four-expert evaluations on a subset of 75 paper--patent pairs assign Zero-Shot the lowest legal compliance rating (1.5 versus 4.7 for \textsc{FlowPlan-G2P}). Under expert-aligned Pat-DEVAL evaluation, \textsc{FlowPlan-G2P} outperforms prompting and outline-based baselines across all four dimensions and lifts open-weight Llama-4-Scout (4.3) above Few-Shot Sonnet-4.5 (2.3). Ablations further show that concept graph induction primarily supports technical fidelity, data precision, and legal compliance, while section-level planning governs structural coverage. Future work includes reducing multi-call inference overhead by distilling concept extraction and section planning into lightweight models, incorporating multimodal figure inputs alongside text, and extending the induced concept graph toward joint claim and description drafting.
\section*{Limitations}

The current pipeline drafts patent descriptions in isolation rather than jointly generating claims or verifying claim-to-description support, which would require coupling description planning with claim-topology decoders~\citep{yoo2026spg}. Within Stage~1 and Stage~2, text previews are truncated to 200--300 characters during graph induction and 150 characters during section planning, and deterministic graph merging consolidates nodes strictly by identifier rather than semantic equivalence, leaving paraphrased duplicates unmerged across candidates. In Stage~2, the entropy-based homogeneity score $Sim_i$ evaluates to zero on evenly balanced multi-category sections and $C_i$ counts only intra-section edges, so balanced canonical plans can rely on the highest-score fallback governed by $P_{\text{pen}}$ and canonical ordering. Because per-document threshold-pass versus fallback counts were not logged during experimentation, we cannot isolate their relative frequencies, and replacing $Sim_i$ with a section-conditioned target distribution metric remains an important refinement. In addition, figure descriptions rely solely on paper text rather than multimodal parsing of technical diagrams~\citep{shukla2025patentlmm}, and the pipeline does not verify novelty against external prior-art corpora or multi-aspect patent similarity graphs~\citep{yoo2025patentmind}.

Our evaluation covers 146 curated English paper--patent pairs from Pap2Pat-EvalGold, together with 30 unfiltered pairs and a subset of 75 paper--patent pairs rated by four experts. These results do not establish generalization to non-English jurisdictions, office-specific statutory conventions outside US and European drafting styles, or informal early-stage invention disclosures. Automated comparisons rely on Pat-DEVAL as a proxy alongside our four-expert subset rather than formal patent office examination, and our baseline comparisons do not equalize total inference calls across single-pass and multi-stage pipelines.

Decomposing each document into nine sequential reasoning calls, two LLM graph-extraction calls, five candidate section plans, and up to five section-generation calls requires 21 LLM API calls per description when all five canonical sections are generated (excluding retries), raising latency and token cost over single-pass prompting. Distilling structural graph inductive biases into lighter or graph-free inference encoders~\citep{yoo2026phage} offers a practical path toward reducing this runtime cost. Finally, while the three-stage architecture is modular, its nine categories, opening boilerplates, and penalty rules are specific to patent descriptions; adapting the framework to clinical or regulatory domains would require domain-agnostic semantic graph representations~\citep{seo2026topic} and domain-specific expert validation.

\section*{Ethical Considerations}

\textsc{FlowPlan-G2P} is intended strictly as a drafting aid for patent practitioners rather than a substitute for qualified legal counsel. Generated specifications may contain technical omissions, unsupported embodiments, or statutory defects, and legal responsibility for accuracy, enablement, and validity remains with qualified human reviewers.

Users must verify that they hold appropriate intellectual property rights to source scientific manuscripts and avoid reproducing third-party proprietary material. Because patentability depends on confidentiality prior to filing, practical deployment should rely on private or local inference infrastructure so that unpublished inventions are neither leaked nor retained for external model training. During the preparation of this manuscript, the authors used generative AI to refine linguistic clarity and assist with diagram preparation, reviewed all outputs, and take full responsibility for the final paper.

%\bibliography{anthology,custom}
\bibliography{custom}
\bibliographystyle{acl_natbib}

\appendix
\section{Implementation Specifications}
\label{app:impl_specs}

Table~\ref{tab:impl_specs} summarizes the exact implementation parameters of \textsc{FlowPlan-G2P} from our public repository (\texttt{flowplan\_g2p}). OpenAI endpoints use temperature $T_{\mathrm{gen}} = 0.2$, nucleus sampling $\text{top-}p = 0.9$, and $\text{max\_tokens} = 4096$, while Anthropic and Bedrock endpoints use $\text{max\_tokens} = 4096$ with provider-default sampling. All LLM calls are wrapped with up to three exponential-backoff retries (base delay $1.0$\,s, skipping authentication errors). Across all three stages, generating one patent description executes 21 LLM API calls when all five canonical sections are generated and retries are excluded (9 sequential reasoning calls, 2 LLM graph-extraction calls, 5 section-planning calls, and 5 section-generation calls).

\begin{table}[t]
\centering
\footnotesize
\setlength{\tabcolsep}{4pt}
\begin{adjustbox}{max width=\columnwidth}
\begin{tabular}{lll}
\toprule
\textbf{Stage} & \textbf{Parameter / Rule} & \textbf{Value} \\
\midrule
OpenAI LLM & Temperature / $\text{top-}p$ & $0.2$ / $0.9$ \\
Global LLM & Max tokens / API retries & $4096$ / $3$ \\
\midrule
Stage~1 & Sequential reasoning calls & $9$ (200-char prior) \\
Stage~1 & Candidate graphs ($G_1, G_2, G_3$) & $1$ rule + $2$ LLM \\
Stage~1 & LLM graph preview length & $300$ chars / node \\
\midrule
Stage~2 & Candidate plans ($k$) & $5$ (150-char preview) \\
Stage~2 & Thresholds ($\tau_C$ / $\tau_S$) & $0.5$ / $0.6$ \\
Stage~2 & Penalty (mand. / pair / order) & $1.0$ / $0.5$ / $0.3$ \\
\midrule
Stage~3 & Section generation calls & up to $5$ (non-empty $S_1\dots S_5$) \\
\bottomrule
\end{tabular}
\end{adjustbox}
\caption{Implementation specifications and stage-wise call budget (21 LLM calls per document when all five canonical sections are generated, excluding retries).}
\label{tab:impl_specs}
\end{table}

\section{Stage~1 Reasoning and Extraction Prompts}
\label{app:expert_prompts}

Table~\ref{tab:category_boilerplates} lists all nine patent drafting categories $\mathcal{S}$ and the exact opening boilerplate phrases used in \texttt{concept\_induction.py}. Each sequential reasoning step injects the full source paper $D$, the leading 200-character prefixes of prior outputs $R_{1:i-1}$, step-specific extraction guidance, and the required opening phrase.

\begin{table}[t]
\centering
\footnotesize
\setlength{\tabcolsep}{3.5pt}
\begin{adjustbox}{max width=\columnwidth}
\begin{tabular}{ll}
\toprule
\textbf{Category} & \textbf{Required Opening Boilerplate} \\
\midrule
\textit{Field} & ``The present invention relates to'' \\
\textit{TechProblem} & ``However, conventional technology has the following drawbacks'' \\
\textit{PriorArt} & ``According to the prior art'' \\
\textit{Novelty} & ``The novelty of the present invention lies in'' \\
\textit{Solution} & ``An object of the present invention is to provide'' \\
\textit{Implementation} & ``According to one embodiment of the present invention'' \\
\textit{Effects} & ``Therefore, according to the present invention, the effect is'' \\
\textit{Embodiment} & ``In a preferred embodiment'' \\
\textit{Figure} & ``Figure 1 illustrates'' \\
\bottomrule
\end{tabular}
\end{adjustbox}
\caption{Nine sequential drafting categories and opening boilerplates used in Stage~1.}
\label{tab:category_boilerplates}
\end{table}

\subsection{Sequential Reasoning Prompt Template}
\label{app:expert_prompt_templates}

\noindent\begin{minipage}{\columnwidth}
\begin{lstlisting}[basicstyle=\ttfamily\scriptsize,breaklines=true,breakatwhitespace=true,columns=fullflexible,frame=single,rulecolor=\color{black!35},backgroundcolor=\color{black!2},xleftmargin=3.5pt,xrightmargin=3.5pt,aboveskip=4pt,belowskip=4pt]
Analyze the following scientific paper for the
patent drafting category: {Category}.

Paper content:
{paper_content}

Previous analysis:
{truncated_prior_outputs_200_chars_each}

Task: {category_specific_guidance}
Format your response starting with:
"{boilerplate}..."
\end{lstlisting}
\end{minipage}

\subsection{LLM Candidate Graph Extraction Prompt}
\label{app:llm_graph_prompt}

Candidates $G_2$ and $G_3$ prompt the LLM over 300-character previews of $\{R_i\}_{i=1}^{9}$ to recover implicit dependencies beyond the seven rule-based edges.

\noindent\begin{minipage}{\columnwidth}
\begin{lstlisting}[basicstyle=\ttfamily\scriptsize,breaklines=true,breakatwhitespace=true,columns=fullflexible,frame=single,rulecolor=\color{black!35},backgroundcolor=\color{black!2},xleftmargin=3.5pt,xrightmargin=3.5pt,aboveskip=4pt,belowskip=4pt]
Construct a directed concept graph from the
reasoning outputs below.

Reasoning outputs (300-char previews):
{category}: {content[:300]} ...

Valid relation types:
solves, implements, causes, improves, validates

Return JSON with "nodes" and "edges":
{
  "nodes": [
    {"id": "problem_1", "type": "TechProblem",
     "content": "..."},
    {"id": "solution_1", "type": "Solution",
     "content": "..."}
  ],
  "edges": [
    {"source": "problem_1", "target": "solution_1",
     "relation": "solves"}
  ]
}
\end{lstlisting}
\end{minipage}

\section{Graph Merging, Repair, and Section Planning}
\label{app:graph_construction}

This section details the deterministic consolidation procedure in Section~\ref{sec:stage1} and the section-assignment configuration in Section~\ref{sec:stage2}.

\paragraph{Rule-Based Candidate ($G_1$).}
For each non-empty reasoning output $R_i$, one node is created with identifier \texttt{\{category\}\_1}. Whenever both endpoints are present, seven canonical directed edges are added, namely $\textit{TechProblem} \xrightarrow{\textit{solves}} \textit{Solution}$, $\textit{Solution} \xrightarrow{\textit{implements}} \textit{Implementation}$, $\textit{Implementation} \xrightarrow{\textit{causes}} \textit{Effects}$, $\textit{Solution} \xrightarrow{\textit{improves}} \textit{Effects}$, $\textit{Embodiment} \xrightarrow{\textit{validates}} \textit{Implementation}$, $\textit{Figure} \xrightarrow{\textit{validates}} \textit{Embodiment}$, and $\textit{Novelty} \xrightarrow{\textit{improves}} \textit{Solution}$. If structured JSON parsing fails for $G_2$ or $G_3$, the pipeline falls back to $G_1$.

\paragraph{Deterministic Merging and Verification.}
To form $G^*$ from $\{G_1, G_2, G_3\}$, \texttt{ConceptGraph.\allowbreak merge\_graphs} aggregates nodes by identifier union across candidates (keeping the first seen node content per identifier) and groups edges by ordered endpoint pair $(u, v)$. When multiple candidates emit conflicting relation types for the same pair $(u, v)$, majority voting selects the most frequent relation type. Post-merge verification (\texttt{\_verify\_graph}) checks that all mandatory categories (\textit{Field}, \textit{TechProblem}, \textit{Solution}) exist, inserting placeholder nodes (\texttt{\{category\}\_\allowbreak placeholder}) when absent, and connects any isolated \textit{Effects} node from a \textit{Solution} node via a \textit{causes} edge.

\paragraph{Section Mapping and Assignment Prompt.}
Table~\ref{tab:section_mapping} reports the five canonical patent sections, default fallback category mappings, and order priorities used in \texttt{paragraph\_planning.py}. In each of the $k = 5$ planning draws, the LLM receives 150-character node previews and all directed edges of $G^*$ and returns a JSON dictionary mapping section names to node identifiers. Any unassigned node or malformed output falls back to Table~\ref{tab:section_mapping}.

\begin{table}[t]
\centering
\footnotesize
\setlength{\tabcolsep}{3.5pt}
\begin{adjustbox}{max width=\columnwidth}
\begin{tabular}{cll}
\toprule
\textbf{Order} & \textbf{Section} & \textbf{Default Categories} \\
\midrule
1 & Field & \textit{Field} \\
2 & Background & \textit{TechProblem}, \textit{PriorArt} \\
3 & Summary & \textit{Solution}, \textit{Novelty} \\
4 & Detailed Description & \textit{Implementation}, \textit{Embodiment}, \textit{Figure} \\
5 & Effects & \textit{Effects} \\
\bottomrule
\end{tabular}
\end{adjustbox}
\caption{Canonical section order and default fallback category mapping in Stage~2.}
\label{tab:section_mapping}
\end{table}

\noindent\begin{minipage}{\columnwidth}
\begin{lstlisting}[basicstyle=\ttfamily\scriptsize,breaklines=true,breakatwhitespace=true,columns=fullflexible,frame=single,rulecolor=\color{black!35},backgroundcolor=\color{black!2},xleftmargin=3.5pt,xrightmargin=3.5pt,aboveskip=4pt,belowskip=4pt]
Assign each node in the concept graph to one
canonical patent section: Field, Background,
Summary, Detailed Description, or Effects.

Nodes (150-char previews):
- {node_id} ({node_type}): {content[:150]}

Edges:
- {source_id} -> {target_id} ({relation})

Return JSON:
{"sections": {"Field": [...], ...}}
\end{lstlisting}
\end{minipage}

\section{Graph-Conditioned Generation Prompts}
\label{appendix:prompt}

In Stage~3 (\texttt{graph\_generation.py}), each ordered section subgraph $S_i = (V_i, E_i)$ is serialized by \texttt{\_linearize\_\allowbreak subgraph} into a concept list and a relation list before prompting the LLM.

\noindent\begin{minipage}{\columnwidth}
\begin{lstlisting}[basicstyle=\ttfamily\scriptsize,breaklines=true,breakatwhitespace=true,columns=fullflexible,frame=single,rulecolor=\color{black!35},backgroundcolor=\color{black!2},xleftmargin=3.5pt,xrightmargin=3.5pt,aboveskip=4pt,belowskip=4pt]
Concepts:
- {node.id} ({node.type}): {node.content}

Relations:
- {edge.source} {edge.relation} {edge.target}
\end{lstlisting}
\end{minipage}

Each section prompt concatenates the shared patent drafting instruction below, the section name and opening boilerplate, the linearized subgraph, and a section-specific strategy block.

\noindent\begin{minipage}{\columnwidth}
\begin{lstlisting}[basicstyle=\ttfamily\scriptsize,breaklines=true,breakatwhitespace=true,columns=fullflexible,frame=single,rulecolor=\color{black!35},backgroundcolor=\color{black!2},xleftmargin=3.5pt,xrightmargin=3.5pt,aboveskip=4pt,belowskip=4pt]
Using the following concepts and relations,
generate the {section_name} section of a patent
Description.

Required Opening: "{section_boilerplate}..."

Subgraph:
{linearized_subgraph}

Shared Requirements:
- Maintain formal patent drafting register
- Elaborate key technical features across
  alternative embodiments where applicable
- Include explicit figure references
  (e.g., "Figure 1 illustrates...")

Section Strategy ({section_name}):
{section_specific_bullet_instructions}
\end{lstlisting}
\end{minipage}

Each section-specific strategy block operationalizes the section roles summarized in Section~\ref{sec:stage3}, and \texttt{\_post\_process\_\allowbreak section} normalizes whitespace and prepends canonical uppercase headings under a unified \texttt{DESCRIPTION} header.

\end{document}